\documentclass[10pt,twocolumn,letterpaper]{article}

\usepackage{wacv}              

\usepackage{graphicx}
\usepackage{amsmath}
\usepackage{amssymb}
\usepackage{booktabs}
\usepackage{float}
\usepackage{stfloats}

\usepackage[pagebackref,breaklinks,colorlinks]{hyperref}

\usepackage[capitalize]{cleveref}
\crefname{section}{Sec.}{Secs.}
\Crefname{section}{Section}{Sections}
\Crefname{table}{Table}{Tables}
\crefname{table}{Tab.}{Tabs.}

\def\wacvPaperID{484} 
\def\confName{WACV}
\def\confYear{2024}

\begin{document}

\title{Hybrid Neural Diffeomorphic Flow for Shape Representation and Generation via Triplane}


\author{Kun Han\textsuperscript{1}\;\;\; 
Shanlin Sun\textsuperscript{1}\;\;\; 
Xiaohui Xie\textsuperscript{1}\\
\textsuperscript{1}University of California, Irvine, USA \;\;\;\;\;\; \\
{\tt\small \{khan7,shanlins,xhx\}@uci.edu} \\
}

\twocolumn[{%
\renewcommand\twocolumn[1][]{#1}%
\maketitle
\begin{center}
    \centering
    \captionsetup{type=figure}
    
    \vspace{-2em}
    \includegraphics[width=0.9\linewidth]{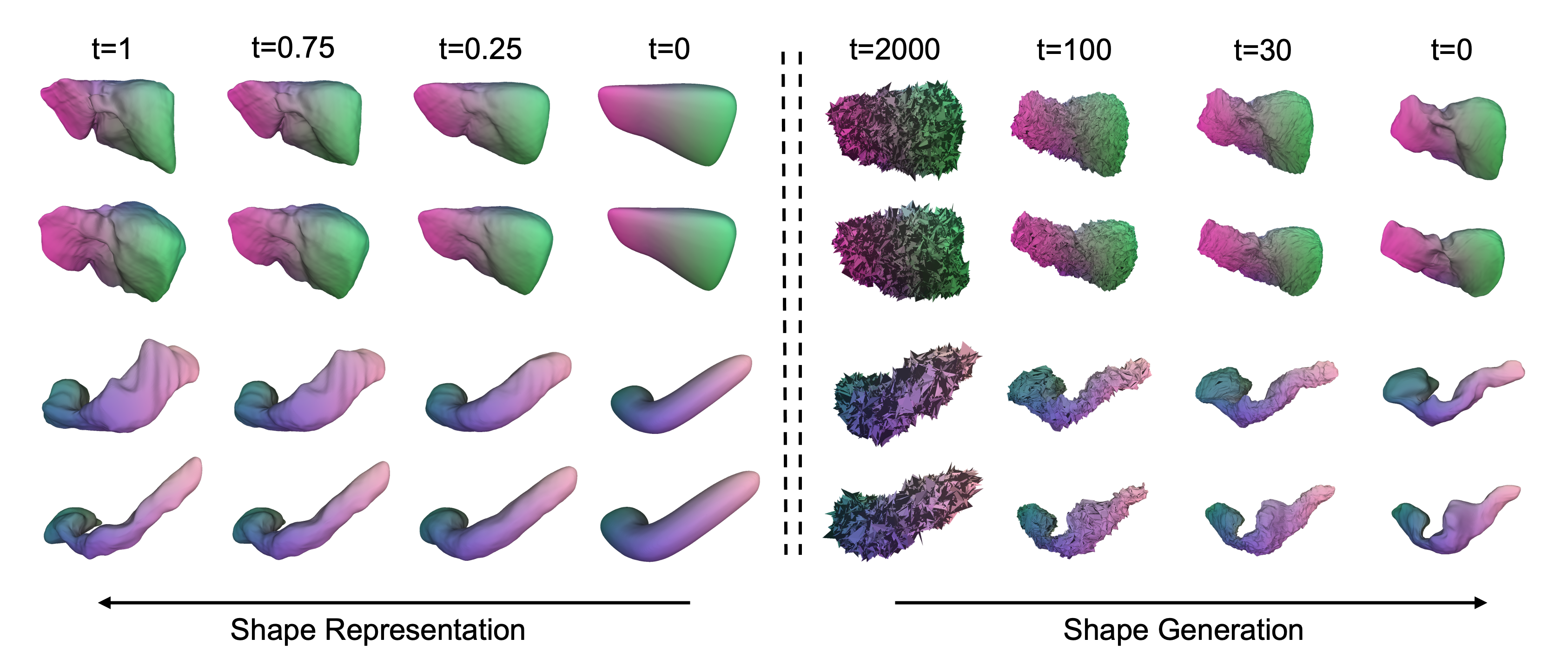}
    \label{fig:vis_all}
    \vspace{-1em}
    \captionof{figure}{\textbf{Shape Representation} and \textbf{Shape Generation}. \textbf{Left} half presents the diffeomorphic deformation from the learned \textbf{template} (t=0) to instance shapes (t=1), with color highlighting the dense correspondence captured by triplane features. \textbf{Right} half presents the denoising process for shape generation. The shapes are generated as deformed templates and the 3D deformation is controlled by the generated triplane features from diffusion.}
\end{center}%
}]


\begin{abstract}
   Deep Implicit Functions (DIFs) have gained popularity in 3D computer vision due to their compactness and continuous representation capabilities. However, addressing dense correspondences and semantic relationships across DIF-encoded shapes remains a critical challenge, limiting their applications in texture transfer and shape analysis. Moreover, recent endeavors in 3D shape generation using DIFs often neglect correspondence and topology preservation. This paper presents HNDF (\underline{H}ybrid \underline{N}eural \underline{D}iffeomorphic \underline{F}low), a method that implicitly learns the underlying representation and decomposes intricate dense correspondences into explicitly axis-aligned triplane features. To avoid suboptimal representations trapped in local minima, we propose hybrid supervision that captures both local and global correspondences. Unlike conventional approaches that directly generate new 3D shapes, we further explore the idea of shape generation with deformed template shape via diffeomorphic flows,  where the deformation is encoded by the generated triplane features. Leveraging a pre-existing 2D diffusion model, we produce high-quality and diverse 3D diffeomorphic flows through generated triplanes features, ensuring topological consistency with the template shape. Extensive experiments on medical image organ segmentation datasets evaluate the effectiveness of HNDF in 3D shape representation and generation.
\end{abstract}

\section{Introduction}
\label{sec:intro}

3D geometry representation is critical for numerous computer vision tasks, including 3D model reconstruction, matching and manipulation. Deep implicit functions (DIFs) have emerged as promising alternatives to traditional representation methods such as voxel grids, point clouds and polygon meshes. DIFs offer several advantages such as compactness, continuity, and the ability to capture fine geometric details. They enable efficient computation while leveraging deep neural networks for end-to-end training, enhancing shape representation and understanding.

However, despite the promising results in direct object modeling using DIFs, it is important to consider the common shape features and semantic correspondences shared among objects. Conventional DIFs face challenges in establishing correspondences between different shapes, limiting their applicability in domains like medical image segmentation \cite{lorenz19993d,heimann2009statistical,lamecker20043d} and texture transfer \cite{mertens2006texture,efros2001image}. Previous methods \cite{zheng2021deep,deng2021deformed,sun2022topology} have proposed shape modeling as conditional deformations of a template DIF to address this limitation. However, these methods still have limitations, such as being topology-agnostic or lacking the capability to capture correspondences for local details.

Recent researches have also explored the integration of DIFs for the 3D shape generation\cite{zheng2022sdf,or2022stylesdf,nam20223d, shue20233d}. Compared to point clouds and polygon meshes, DIF-based generation offers continuous representations with high quality and resolution. However, existing approaches primarily focus on direct shape generation without considering underlying point correspondence and topology preservation.

To overcome these challenges, we introduce \underline{H}ybrid \underline{N}eural \underline{D}iffeomorphic \underline{F}low (HNDF) for shape representation and generation. HNDF models shapes as conditional deformations of a template DIF, similar to previous work \cite{zheng2021deep,deng2021deformed,sun2022topology,yang2022implicitatlas}. However, HNDF encodes diffeomorphic deformations into axis-aligned triplane features to enhance representation capability. Local deformations are controlled through interpolation of triplane features with a shared feature decoder. Nevertheless, the direct application of triplanes may lead to local optimization issues and defective deformations, resulting in inaccurate representations. To address this, we propose a hybrid supervision approach that considers both local and global correspondences, along with additional modifications and regularization to preserve the diffeomorphism property of the represented deformations. This combination of triplane feature exploration and supervision enables high representation capabilities and accurate dense correspondences.

Unlike conventional 3D shape generation works which primarily focus on direct shape generation, we explore the idea of deformation-based shape generation, where the template shape is deformed based on newly generated diffeomorphic deformations. This approach ensures that the newly generated shapes maintain the same topology as the template shape, preserving topological consistency while offering a wide range of diverse shapes. To achieve this, we represent deformations using optimized per-object triplane features, which encode diffeomorphic deformations as three axis-aligned 2D feature planes. We concatenate the triplane features as multi-channel images and leverage the existing 2D diffusion models to generate new triplane features. By applying the new diffeomorphic deformations encoded in the triplane features, we deform the template shape to generate novel 3D shapes while preserving their topological characteristics.

The contributions of this paper are as follows:
\begin{enumerate}
    \item We propose HNDF, which leverages axis-aligned triplane features to provide high representation capability and capture dense correspondences accurately.
    \item We demonstrate that hybrid supervision and regularization are essential for ensuring correct deformation representation and preventing the representation from local optima.
    \item Rather than directly generating 3D shapes, we explore the concept of shape generation through diffeomorphic deformations and provide a baseline method utilizing 2D diffusion model. The topology and correspondences are preserved in newly generated 3D shapes.
\end{enumerate}

\section{Related Works}

\noindent\textbf{Deep Implicit Function} Deep implicit functions, or neural fields, have enabled the parameterization of physical properties and dynamics through simple neural networks~\cite{xie2022neural,park2019deepsdf,mescheder2019occupancy,chen2019learning,sitzmann2019scene,mildenhall2021nerf}.
DeepSDF \cite{park2019deepsdf} serves as an auto-decoder model, commonly used as a baseline for shape representation \cite{chabra2020deep,jiang2020local,tretschk2020patchnets}. 
NeRF \cite{park2019deepsdf} presents a novel approach for synthesizing photorealistic 3D scenes from 2D images. 
Occupancy Network \cite{mescheder2019occupancy} constructs solid meshes through the classification of 3D points, while Occupancy Flow \cite{niemeyer2019occupancy} extends this idea to 4D with a continuous vector field in time and space. Recent trends incorporate locally conditioned representations \cite{chabra2020deep,jiang2020local,tretschk2020patchnets, chen2019learning, reiser2021kilonerf}, utilizing small MLPs that are computationally and memory-efficient while capturing local details effectively. One such representation is the hybrid triplane \cite{peng2020convolutional, liu2020neural, martel2021acorn, devries2021unconstrained, chan2022efficient}, which represents features on axis-aligned planes and aggregates them using a lightweight implicit feature decoder. In our work, we adopt the expressive triplane representation. However, instead of decoding the 3D object itself, we utilize triplane features to decode complex diffeomorphic deformations, allowing us to represent new 3D objects by deforming the template shape using the encoded deformation.

\noindent\textbf{Point Correspondence and Topology preservation} Capturing dense correspondences between shapes remains a significant challenge and a critical area of interest in the 3D vision community. Various approaches have been proposed to address point correspondence, including template learning, elementary representation, and deformation field-based methods. Among them, mesh-based methods \cite{litany2017deep,lim2018simple} face difficulties in handling topological changes, sensitivity to mesh connectivity, and challenges in capturing fine-grained details. Elementary-based methods \cite{groueix2018papier,jiang2020local}, on the other hand, may struggle with capturing high-level structural features due to the simplicity of the elements used. DIT \cite{zheng2021deep} and NDF \cite{sun2022topology} exemplify deformation field-based methods, with DIT exhibiting smoother deformations using LSTM \cite{hochreiter1997long} and NDF employing NODE \cite{chen2018neural} for achieving diffeomorphic deformation. ImplicitAtlas \cite{yang2022implicitatlas} integrates multiple templates to improve the shape representation capacity at a negligible computational cost. In our work, we follow the NDF framework but enhance the representation's capacity to capture accurate correspondences by leveraging more powerful triplane representation. Experimental results highlight the importance of incorporating triplane features with hybrid supervision, which prevents local optimization issues, provides significantly more accurate correspondences, and ensures the preservation of topology.

\noindent\textbf{3D Shape Generation} Generative models, such as GANs, autoregressive models, score matching models, and denoising diffusion probabilistic models, have been extensively studied for 3D shape generation. However, GAN-based methods \cite{chan2022efficient,gao2022get3d,gadelha20173d,gu2021stylenerf,meng2021gnerf,nguyen2019hologan,or2022stylesdf} still outperform alternative approaches. Voxel-based GANs \cite{gadelha20173d,henzler2019escaping,wu2016learning}, for example, directly extend the use of CNN generators from 2D to 3D settings with high memory requirement and computational burden. In recent years, there has been a shift towards leveraging expressive 2D generator backbones, such as StyleGAN2 \cite{karras2020analyzing}. EG3D \cite{chan2022efficient} combines a hybrid explicit-implicit triplane representation to improve computational efficiency while maintaining expressiveness. Get3D\cite{gao2022get3d} incorporates the deformable tetrahedral grid for explicit surface extraction and triplane representation for differentiable rendering to generate textured 3D shapes. 

Compared to the existing GAN-based approaches for 3D generation, the development of 3D diffusion models is still in its early stages. Several notable works have explored the application of diffusion models in generating 3D shapes. PVD \cite{zhou20213d} proposed the use of a point-voxel representation combined with PVConv\cite{liu2019point} to generate 3D shapes through diffusion. DPM \cite{luo2021diffusion} introduced a shape latent code to guide the Markov chain in the reverse diffusion process. 
MeshDiffusion \cite{liu2023meshdiffusion} utilized the deformable tetrahedral grid parametrization for unconditionally generating 3D meshes. 3D-LDM \cite{nam20223d} integrated DeepSDF\cite{park2019deepsdf} into diffusion-based shape generation, leveraging diffusion to generate a global latent code and improve the conditioning of the neural field. NFD \cite{shue20233d} extended the use of 2D diffusion into 3D shape generation, exploring the potential of diffusion models in capturing and generating complex 3D shapes with Occupancy Network \cite{mescheder2019occupancy}.

While existing approaches in shape generation focus on directly generating 3D shapes, they often neglect the preservation of underlying topology. This oversight can lead to artifacts in the generated shapes and limit their applicability in scenarios where topology is important. In our work, we introduce a baseline diffusion-based method that deforms a template to generate new shape. The diffeomorphic deformation is encoded by the generated triplane features. Our approach focuses on producing visually coherent and realistic shapes while preserving point correspondence and underlying topology.

\section{Preliminaries}
\label{sec:prelim}

\noindent\textbf{Diffeomorphic Flow} is a continuous and smooth mapping that transforms a given manifold or space while preserving its differentiable structure. In the context of 3D geometry, diffeomorphic flow plays a crucial role in establishing dense point correspondences between 3D shapes and ensuring the preservation of their underlying topology during deformation. Mathematically, the forward diffeomorphic flow  $\Phi(p, t): \mathbb{R}^3 \times[0,1] \rightarrow \mathbb{R}^3$ describes the trajectory of a 3D point $p$ over the interval $[0,1]$, where the starting point $p$ is located in the space of instance shape $S$ and the destination point corresponds to the target shape $T$. The velocity field $\mathbf{v}(p, t): \mathbb{R}^3 \times[0,1] \rightarrow \mathbb{R}^3$ represents the derivative of deformation of 3D points. The diffeomorphic flow $\Phi$ is obtained by solving the initial value problem (IVP) of an ordinary differential equation (ODE),
\vspace{-0.7em}
\begin{equation}
\label{eq:ode_forward}
    \frac{\partial \Phi}{\partial t}(p, t)=\mathbf{v}\left(\Phi(p, t), t\right) \quad \text { s.t. } \quad \Phi(p, 0)=p
    \vspace{-0.7em}
\end{equation}

Similarly, the inverse flow $\Psi$ can be calculated by solving a corresponding ODE with negative velocity field $-\mathbf{v}$, allowing for the transformation from the template space to the instance space
\vspace{-0.7em}
\begin{equation}
\label{eq:ode_backward}
    \frac{\partial \Psi}{\partial t}(p, t)=-\mathbf{v}\left(\Psi(p, t), t\right) \quad \text { s.t. } \quad \Psi(p, 0)=p
    \vspace{-0.7em}
\end{equation}
where $p$ is the starting point on the target shape. The property of topology preservation is achieved through the Lipschitz continuity of the velocity field. The forward and backward diffeomorphic deformation can be calculated by the integration of the velocity field by solving the equation \ref{eq:ode_forward} \ref{eq:ode_backward}, respectively.

\vspace{1em}

\noindent\textbf{Diffusion Probabilistic Model} (DPM)\cite{ho2020denoising} is a parameterized Markov chain designed to learn the underlying data distribution $p(X)$. 

During the Forward Diffusion Process (FDP), the diffused data point $X_t$ is obtained at each time step t by sampling from the conditional distribution:
\vspace{-0.7em}
\begin{equation}
    q\left(X_t \mid X_{t-1}\right)=\mathcal{N}\left(X_t; \sqrt{1-\beta_t} X_{t-1}, \beta_t I\right)
    \vspace{-0.7em}
\end{equation}
where $X_0$ is sampled from the initial distribution $q(X_0)$, and $X_T$ follows a Gaussian distribution $N(X_T ; 0, I)$. The parameter $\beta_t \in(0,1)$ represents a variance schedule that gradually introduces Gaussian noise to the data. By defining $\alpha_t=1-\beta_t$ and $\bar{\alpha}_t=\prod_{s=1}^t\left(1-\beta_t\right)$, $X_t$ can be sampled conditionally on $X_0$ as $q\left(X_t \mid X_0\right)=\mathcal{N}\left(X_t ; \sqrt{\bar{\alpha_t}} X_0,\left(1-\bar{\alpha_t}\right) I\right)$, providing a distribution for sampling $X_t$ from the initial data $X_0$.

In contrast, the Reverse Diffusion Process aims to approximate the posterior distribution $p(X_{t-1} | X_t)$ to recreate a realistic $X_0$ starting from random noise $X_T$. The Reverse Diffusion Process is formulated as a trajectory of posterior distributions starting from $X_T$:
\vspace{-0.7em}
\begin{equation}
    p\left(X_{0: T}\right)=p\left(X_T\right) \prod_{t=1}^T p_\theta\left(X_{t-1} \mid X_t\right)
    \vspace{-0.7em}
\end{equation}
The conditional distribution $p_\theta(X_{t-1} | X_t)$ is approximated by a neural network with parameters $\theta$:
\vspace{-0.7em}
\begin{equation}
    p_\theta\left(X_{t-1} \mid X_t\right)=\mathcal{N}\left(X_t ; \mu_\theta\left(X_t, t\right), \Sigma_\theta\left(X_t, t\right)\right)
    \vspace{-0.7em}
\end{equation}


\vspace{0.3em}
\section{Method}

In this section, we present our Hybrid Neural Diffeomorphic Flow (HNDF) for shape representation and generation. Section \ref{subsec:review_ndf} reviews our baseline method \cite{sun2022topology}. In Section \ref{subsec:endf}, we introduce the utilization of triplane features, and the hybrid supervision for capturing local and global correspondences. Finally, in Section \ref{subsec:shape_gen}, we describe our proposed method for generating topology-preserving shapes.

\subsection{Review of NDF}
\label{subsec:review_ndf}

NDF \cite{sun2022topology}, similar to DeepSDF\cite{park2019deepsdf}, represents a 3D shape $S_i$ using a continuous signed distance field (SDF) $\mathcal{F}$. Given a random 3D point $p$ and a one-dimensional latent code $c_i$ of length $k$, $\mathcal{F}$ outputs the distance from the point $p$ to the closest surface of shape $S_i$. However, unlike DeepSDF, which directly represents 3D shapes, NDF uses a deform code $c_i$ to control the deformation of each instance shape from the template shape. As a result, the conditional continuous SDF $\mathcal{F}$ can be decomposed into $\mathcal{T} \circ \mathcal{D}$, where $\mathcal{D}: \mathbb{R}^3 \times \mathbb{R}^k \mapsto \mathbb{R}^3$ provides the deformation mapping from the coordinates of $p$ in the instance space of $S_i$ to a canonical position $p'$ in the template space. The function $\mathcal{T}$ represents a single shape DeepSDF that models the implicit template shape. 

\subsection{Hybrid Shape Representation via Triplane}
\label{subsec:endf}

\begin{figure}[ht]
\includegraphics[width=\linewidth]{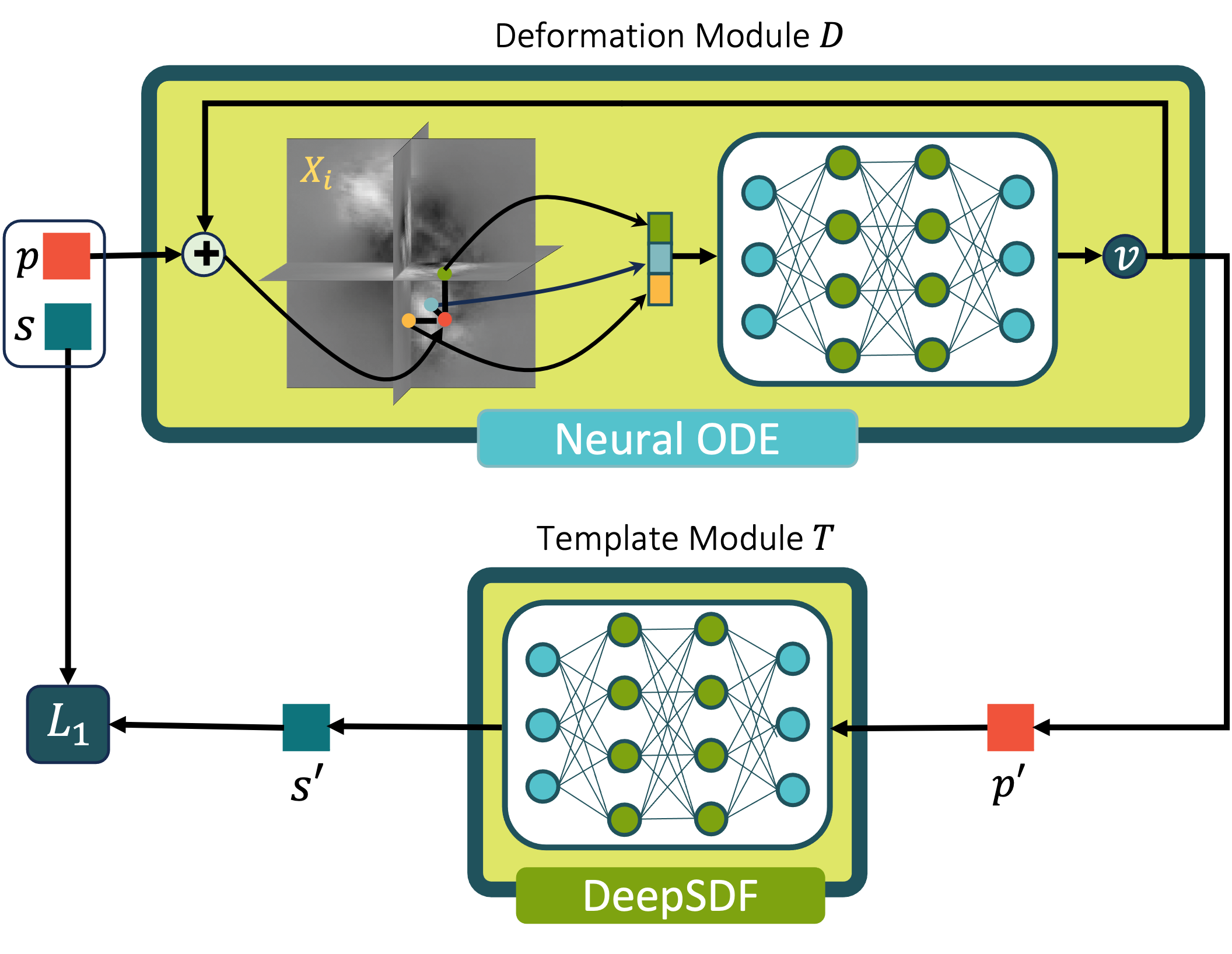}
\vspace{-2em}

\caption{\textbf{Shape Representation} framework consists of a deformation module $\mathcal{D}$, a template module $\mathcal{T}$, and per-object triplane features $X_i$. Given a point $p$ in the instance space, we compute its corresponding destination point $p'$ in the template space using Eq. \ref{eq:ode_forward}. The template module then provides the sign distance value $s'$ for this point. During training, we optimize the framework by minimizing the $L_1$ loss between the represented $s'$ and the ground truth $s$, while incorporating regularization terms.}  
\label{fig:model_rep}
\vspace{-1.5em}
\end{figure}

As shown in \cite{chabra2020deep,jiang2020local,tretschk2020patchnets,peng2020convolutional,chen2019learning, reiser2021kilonerf}, previous methods \cite{park2019deepsdf, zheng2021deep, sun2022topology, yang2022implicitatlas} utilizing a single latent vector to control the entire shape or deformation space could not be able to capture the details of the complex 3D shape or the deformation. Motivated by recent advancements in hybrid representation \cite{chan2022efficient}, we propose to encode complex diffeomorphic deformations as a set of three axis-aligned 2D feature planes, as shown in Fig. \ref{fig:model_rep}. This enables us to capture fine-grained details and variations in the shape space more effectively.

The triplane representation is a hybrid architecture for neural fields that combines explicit and implicit components \cite{chan2022efficient}. For each instance shape $S_i$, it employs three axis-aligned orthogonal feature planes $(X_i = [F_{x y}^i, F_{x z}^i, F_{y z}^i])$, each with a resolution of $L \times L \times C$. These planes serve as the encoded representations of the deformation. To query a deformation, the position of given point $p_i$ is projected onto each of the feature planes, and the corresponding feature vectors are retrieved using bilinear interpolation. Subsequently, a lightweight multilayer perceptron (MLP) decoder is employed to interpret the aggregated features as corresponding velocity vector $v_i$. The diffeomorphic deformation $d_i$ for point $p_i$ can be calculated by integrating the velocity vector using an explicit Runge-Kutta solver \cite{chen2018neural}, as defined in Eq. \ref{eq:ode_forward}. In contrast to the approach in \cite{chan2022efficient}, where feature aggregation is performed through summation, we have found that concatenating the interpolated features from the triplane yields better results. 

\vspace{-0.5em}
\subsubsection{Training}
\label{subsubsec:training}

In our method, we represent the instance shape $S_i$ as a deformed template shape ($\mathcal{T} \circ \mathcal{D}_i$). To capture the continuous shape of $S_i$, we employ two modules: a continuous diffeomorphic deformation module $\mathcal{D}$ and a template shape representation $\mathcal{T}$. As discussed in Sec. \ref{subsec:endf}, the diffeomorphic deformation $d_i$ of a point $p_i$ is obtained by integrating the velocity field. The signed distance field (SDF) value of $p_i$ is determined by evaluating the implicit template shape module $\mathcal{T}$ at the transformed point $p_i'$, where $p_i' = p_i + d_i$. 

During training, our method jointly optimizes the deformation module $\mathcal{D}$, template DeepSDF shape $\mathcal{T}$, and per-object triplane features $X_i$ to represent a training set of $S$ objects. The triplane representation provides an expressive representation power, allowing us to achieve accurate deformation and correspondence. Unlike NDF \cite{sun2022topology}, which requires multiple deformation modules, our method only requires one deformation module. This not only enables more accurate deformation representation but also reduces the memory and computation requirements.

The training objective function includes a reconstruction loss and a regularization loss:
\vspace{-0.7em}
\begin{equation}
\label{loss:train}
    \mathcal{L}_{train}=\mathcal{L}_{\text {rec }}+\lambda_{\text {reg }} \mathcal{L}_{\text {reg }}
    \vspace{-0.7em}
\end{equation}
where $\mathcal{L}_{\text {rec }}$ shows the reconstruction loss between the ground truth SDF value $s_i$ and the represented SDF value $s_i'$, and $\mathcal{L}_{\text {reg }}$ includes a series of regularization terms. Specifically, reconstruction loss $\mathcal{L}_{\text {rec }}$ can be written as
\vspace{-0.7em}
\begin{equation}
\label{loss:rec}
    \mathcal{L}_{\text {rec }}= \sum_{i=1}^S \sum_{j=1}^N L_1 (\mathcal{T} \circ \mathcal{D}_i(p_{i,j}), s_{i, j})
    \vspace{-0.7em}
\end{equation}
where $S$ is the number of instance shapes in the training set, $N$ is the number of sampling points for each shape, $p_{i,j}$ is the $j$-th point on the $i$-th shape and $s_{i, j}$ is the corresponding ground truth SDF value. 

In addition to the point-wise deformation regularization ($\sum_{i,j} \left\|\mathcal{T} \circ \mathcal{D}_i(p_{i,j}) - s_{i, j}\right\|_2$) and the $L_2$ norm feature regularization ($\left\|F_{x y}^{i}\right\|_2+\left\|F_{y z}^{i}\right\|_2+\left\|F_{x z}^{i}\right\|_2$), the inclusion of total variation (TV) regularization \cite{rudin1994total} is crucial for simplifying the triplane representation and ensuring smooth deformations. 
The overall regularization term in the training objective is defined as:
\vspace{-0.7em}
\begin{equation}
\label{loss:reg}
    \mathcal{L}_{\text {reg }} = \lambda_{\text {PW }} \mathcal{L}_{\text {PW }} + \lambda_{\text {L2 }} \mathcal{L}_{\text {L2 }} + \lambda_{\text {TV }} \mathcal{L}_{\text {TV }}
    \vspace{-0.7em}
\end{equation}

\subsubsection{Hybrid Supervision for Inference Time Reconstruction}
\label{subsubsec:rec}

\begin{figure}[ht]
\centering
\vspace{-2em}
\includegraphics[width=0.8\linewidth]{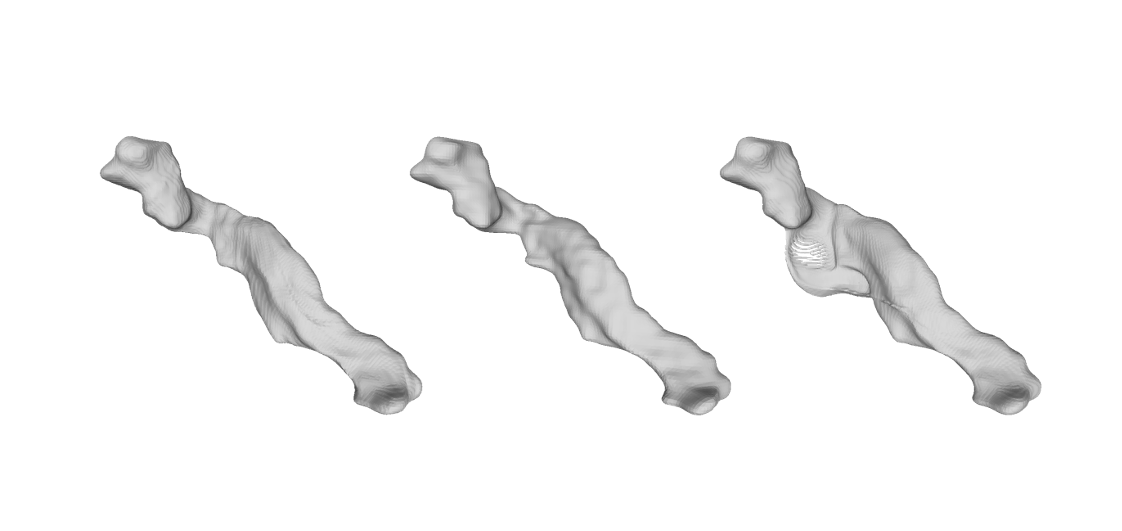}
\vspace{-2em}
\caption{Left is the reconstruction results with proposed hybrid supervision. Middle is the ground truth. Right is the result from purely local supervision, which failed to capture the global correspondence.}  \label{fig:hybrid_sup}
\vspace{-1em}
\end{figure}

In contrast to previous methods \cite{park2019deepsdf,sun2022topology} that utilize a single latent vector for shape reconstruction, the incorporation of triplane representation in our work introduces specific challenges when reconstructing new shapes. 
Specifically, during the optimization process, the features interpolated from the triplane representation for different positions $p_i$ are optimized locally. Since the final diffeomorphic deformation is the integration of velocity vectors along the trajectory in the entire space, the optimized deformation can become trapped in local optima, leading to incorrect global correspondence, as shown in Fig. \ref{fig:hybrid_sup}. As a consequence, the reconstructed shape and deformation may exhibit artifacts, and the overall correspondence may be compromised. 

Therefore, we introduce a hybrid supervision strategy that incorporates both global and local correspondence. In addition to randomly sampled points that provide local supervision, we downsample the entire $N \times N \times N$ coordinate grid with predefined step size and include these regularly sampled points for global supervision during optimization. 

The reconstruction loss during inference is defined as:
\vspace{-0.7em}
\begin{equation}
\mathcal{L}_{\text {rec}} = \mathcal{L}_{\text {rec}}^{\text {grid}} + \lambda_{\text {random }}\mathcal{L}_{\text {rec}}^{\text {random}}
\vspace{-0.7em}
\end{equation}
where $\lambda_{\text {random }}$ is initialized as 0 and gets increased as the optimization continues. 

After we get the grid-structure deformation $\Phi$, we utilize two additional regularization terms to ensure the diffeomorphism of the deformation field and maintain structural integrity. The first term, selective Jacobian determinant regularization ($\mathcal{L}_{\text{Jdet}}$), enforces local orientation consistency. 
\vspace{-0.7em}
\begin{equation}
    \mathcal{L}_{J d e t}=\frac{1}{N} \sum_{p} relu\left(-\left|J_{\Phi}(p)\right|\right)
    \vspace{-0.7em}
\end{equation}
where the Jacobian matrix $J_\Phi$ is defined as:
\begin{equation}
    J_{\Phi}(p)=\begin{bmatrix}
\frac{\partial \Phi_{x}(p)}{\partial x} & \frac{\partial \Phi_{x}(p)}{\partial y} & \frac{\partial \Phi_{x}(p)}{\partial z} \\
\frac{\partial \Phi_{y}(p)}{\partial x} & \frac{\partial \Phi_{y}(p)}{\partial y} & \frac{\partial \Phi_{y}(p)}{\partial z} \\
\frac{\partial \Phi_{z}(p)}{\partial x} & \frac{\partial \Phi_{z}(p)}{\partial y} & \frac{\partial \Phi_{z}(p)}{\partial z}
\end{bmatrix}
\end{equation}

The second term, deformation regularization ($\mathcal{L}_{\text{def}}$), discourages excessively skewed deformations that may lead to unnatural shapes. 
\vspace{-0.7em}
\begin{equation}
    \mathcal{L}_{def}= \sum_{p}\|\nabla \Phi(p)\|^{2}
    \vspace{-0.7em}
\end{equation}

The combination of global and local supervision provides comprehensive guidance during optimization, enabling the model to capture both fine-grained details and global structural consistency.

\subsubsection{Point Correspondence and Shape Registration}
\label{subsubsec:pc_and_sr}

During inference, our method utilizes the learned template shape from training and the diffeomorphic deformation encoded by the triplane feature to establish point correspondence and shape registration between different instance shapes. For each point $p_t$ on the template shape, we apply the inverse diffeomorphic flow $\Psi$, as defined in Eq. \ref{eq:ode_backward}, to obtain the corresponding points $p_i$ and $p_j$ on instance shapes $S_i$ and $S_j$ respectively, based on their respective triplane features $X_i$ and $X_j$. This process allows us to accurately capture point correspondence and establish registration between the instances, facilitating tasks such as shape comparison, shape synthesis, and texture transfer. 

\subsection{Topology-preserving Shape Generation}
\label{subsec:shape_gen}

In this section, we present our proposed method for topology-preserving shape generation. Rather than directly generating shapes from scratch, our approach focuses on generating new shapes by deforming a template shape using synthesized diffeomorphic deformations. 

\subsubsection{Training a Diffusion Model}
\label{subsec:train_dif}

After the training of the diffeomorphic deformation module $\mathcal{D}$ and the template shape representation $\mathcal{T}$, as described in Section \ref{subsubsec:training}, we can leverage the hybrid supervision introduced in Section \ref{subsubsec:rec} to obtain the corresponding per-shape triplane features for the dataset. These optimized sets of triplane features, denoted as $X \in \mathbb{R}^{N \times (L \times L \times 3C)}$, will be utilized to train our generative model, where N denotes the number of shapes in the dataset, L is the dimension of triplane features and C is the number of channels for each 2D plane ($F_{x y}^i, F_{x z}^i, F_{y z}^i$).

\begin{figure}[H]
\vspace{-1.0em}
\includegraphics[width=\linewidth]{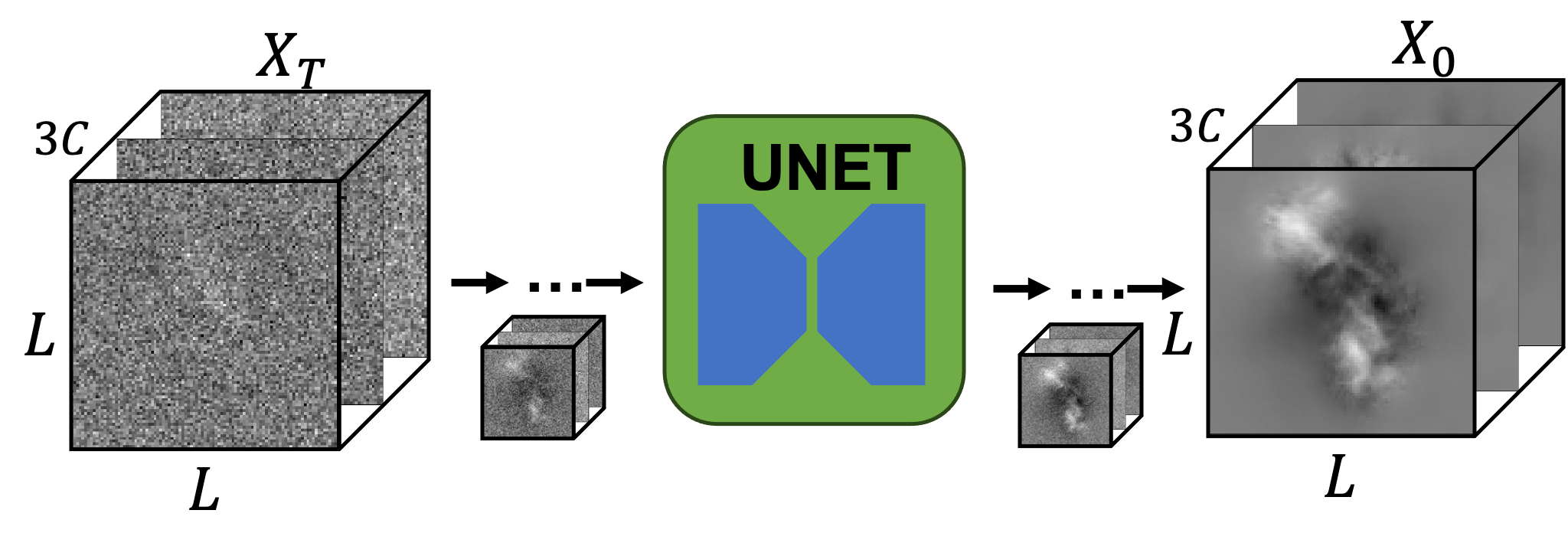}
\vspace{-2em}
\caption{The triplane feature can be represented as multi-channel images. In our work, we adopt the 2D diffusion model as our shape generation model. The generated triplane feature encodes the diffeomorphic deformation that deforms the template to produce the new shapes.}  \label{fig:model_gen}
\vspace{-1em}
\end{figure}

In our framework, the triplane feature is composed of three 2D plane features. We concatenate these feature planes and takes advantage of the strong generative capability of existing 2D diffusion models. Following Sec. \ref{sec:prelim}, we train a diffusion model to learn the reverse diffusion process and predict the added noise from its noisy input by minimizing the following loss function:
\vspace{-0.7em}
\begin{equation}
\begin{aligned}
    Loss(\theta)=& \mathbb{E}_{X_0 \sim q\left(X\right), \epsilon \sim \mathcal{N}(0, I), t} \\
    & \left[\left\|\epsilon-\epsilon_\theta\left(\sqrt{\bar{\alpha}_t} X_0+\sqrt{1-\bar{\alpha}_t} \epsilon, t\right)\right\|^2\right]
    \vspace{-2.7em}
\end{aligned}
\end{equation}
where $\epsilon_\theta$ is predicted noise and $\theta$  represents the model parameters.
\vspace{-0.7em}
\subsubsection{New Shape Generation}

During the inference phase, the generation of a new shape involves deforming the template shape based on the diffeomorphic deformation encoded by the sampled triplane features. Following \cite{ho2020denoising}, we initiate the process by sampling a random Gaussian noise $X_T \sim \mathcal{N}(0, I) \in \mathbb{R}^{L \times L \times 3C}$. Subsequently, we perform iterative denoising for a total of $T$ steps as:
\vspace{-1.4em}
\begin{equation}
    X_{t-1}=\frac{1}{\sqrt{\alpha_t}}\left(X_t-\frac{1-\alpha_t}{\sqrt{1-\bar{\alpha}_t}} \boldsymbol{\epsilon}_\theta\left(X_t, t\right)\right)+\sigma_t \boldsymbol{\epsilon}
    \vspace{-0.7em}
\end{equation}
where $\epsilon \sim \mathcal{N}(0, I)$ if $t > 1$, else, $\epsilon = 0$. 

After sampling, the concatenated triplane feature is split into three axis-aligned 2D planes ($F_{xy}^i, F_{xz}^i, F_{yz}^i$). This generated triplane feature can be interpreted as the diffeomorphic deformation. By following the trajectory defined by the ODE function in Eq. \ref{eq:ode_backward}, each point on the template shape is displaced towards its corresponding destination point in the instance space. Consequently, the new generated shape, known as the deformed template, retains the same underlying topology as the template shape, ensuring consistent connectivity. 

\section{Experiments}
In this section, we present the experiments conducted to evaluate our proposed Hybrid Neural Diffeomorphic Flow (HNDF) for shape \textbf{representation} and \textbf{generation} tasks.

\noindent\textbf{Datasets:} To assess the effectiveness of our shape representation, we utilize the same medical datasets as \cite{sun2022topology}: Pancreas CT \cite{roth2015deeporgan} and Inhouse Liver \cite{chen2021deep}, as these datasets exhibit clear common topology while demonstrating shape variation, making them suitable for our evaluation. For shape generation evaluation, we employ the Abdomen1k dataset \cite{ma2021abdomenct}, consisting of 573 valid liver data and 693 pancreas data after preprocessing and filtering. Please refer to the supplementary material for detailed data sources and preprocessing information.

\noindent\textbf{Shape Representation Evaluation:} We evaluate HNDF for shape representation through two experiments. First, we demonstrate the expressive power of triplane representation and the importance of our hybrid supervision. Evaluation metrics include Chamfer distance (CD) and normal consistency (NC). Second, we evaluate point correspondence and shape registration accuracy, incorporating self-intersection (SI) as an additional metric for geometrical fidelity.

\noindent\textbf{Shape Generation Evaluation:} For shape generation evaluation, following \cite{shue20233d}, we adopt an adapted version of Frechet inception distance (FID). This metric considers rendered shading images of our generated meshes, taking human perception into account. As discussed in \cite{zheng2022sdf}, shading-image FID overcomes limitations of other mesh-based evaluation metrics. FID is computed across 20 views and averaged to obtain a final score
\vspace{-0.7em}
\begin{equation}
    \mathrm{FID}=\frac{1}{20}\left[\sum_{i=1}^{20}\left\|\mu_g^i-\mu_r^i\right\|^2+\operatorname{Tr}\left(\Sigma_g^i+\Sigma_r^i-2\left(\Sigma_r^i \Sigma_g^i\right)^{\frac{1}{2}}\right)\right]
\end{equation}
Additionally, precision and recall scores are reported using the method proposed by \cite{sajjadi2018assessing}. Precision reflects the quality of the rendered images, while recall measures the diversity of the generative model.

\noindent\textbf{Baseline Methods} 
We compare our proposed Hybrid Neural Diffeomorphic Flow (HNDF) with several baselines for the shape representation task. This includes DIT\cite{zheng2021deep}, DIF-Net\cite{deng2021deformed}, and NDF\cite{sun2022topology}, which share the same representation formula as ours, where the shape is represented as a deformed template. We also include AtlasNet\cite{groueix2018papier}, which uses explicit mesh parameterization for shape reconstruction. Additionally, we compare with DeepSDF\cite{park2019deepsdf} and NFD \cite{shue20233d}, which directly represent 3D shapes from scratch.

For the shape generation task, we explore different sampling strategies and generative models. We compare against DeepSDF\cite{park2019deepsdf} and NDF\cite{sun2022topology}, which assume a Gaussian distribution for the global latent vector. We sample new shapes by randomly sampling global vectors from a Gaussian distribution or performing PCA analysis on optimized global latent vectors. We also compare with recent generative models such as point-cloud-based PVD\cite{zhou20213d}, and neural-field-based 3D-LDM \cite{nam20223d} and NFD \cite{shue20233d}. However, it's important to note that these models do not consider the preservation of underlying topology.

\subsection{Shape Representation}

\renewcommand\arraystretch{0.83}
\begin{table*}[ht]
\centering
\begin{center}
\begin{tabular}{lcccccccc}
\toprule
& \multicolumn{4}{c}{Representation}  & \multicolumn{4}{c}{Reconstruction} \\
\cmidrule(lr){2-5} \cmidrule(lr){6-9}
& \multicolumn{2}{c}{CD Mean($\downarrow$)}  & \multicolumn{2}{c}{NC Mean($\uparrow$)} & \multicolumn{2}{c}{CD Mean($\downarrow$)}  
& \multicolumn{2}{c}{NC Mean($\uparrow$)} \\
\cmidrule(lr){2-3} \cmidrule(lr){4-5} \cmidrule(lr){6-7} \cmidrule(lr){8-9}

Model/Data & Pancreas & Liver & Pancreas & Liver & Pancreas & Liver & Pancreas & Liver \\
\midrule
DeepSDF & 0.342 & 0.232 & 0.927 & 0.876 & 0.711 & 0.539 & 0.898 & 0.866 \\ 
NFD & 0.200 & 0.168 & 0.969 & 0.884 & \textbf{0.080} & 0.118 & \textbf{0.982} & \textbf{0.898} \\ 
\midrule
AtlasNet & 4.5 & 1.76 & 0.733 & 0.836 & 8.08 & 3.46 & 0.703 & 0.823\\ 
DIT & 0.349 & 0.303 & 0.929 & 0.878 & 0.63 & 0.509 & 0.903 & 0.87 \\ 
DIF-Net & 0.568 & \textbf{0.122} & \textbf{0.979} & \textbf{0.894} & 4.18 & 1.58 & 0.756 & 0.832 \\ 
NDF & 0.315 & 0.291 & 0.933 & 0.883 & 0.512 & 0.476 & 0.917 & 0.873 \\ 
\midrule
Ours & \textbf{0.133} & 0.266 & 0.965 & 0.889 & 0.082 & \textbf{0.116} & 0.961 & 0.885 \\

\bottomrule

\end{tabular}
\end{center}

\vspace{-1.5em}

\caption{
\textbf{Shape Representation} results on Training Shapes and \textbf{Shape Reconstruction} results on Unseen Shapes. The chamfer distance results shown above are multiplied by $10^3$. 
}
\vspace{-1.5em}
\label{tab:rep}
\end{table*}

\begin{figure}[ht]
\vspace{-1em}
\includegraphics[width=1\linewidth]{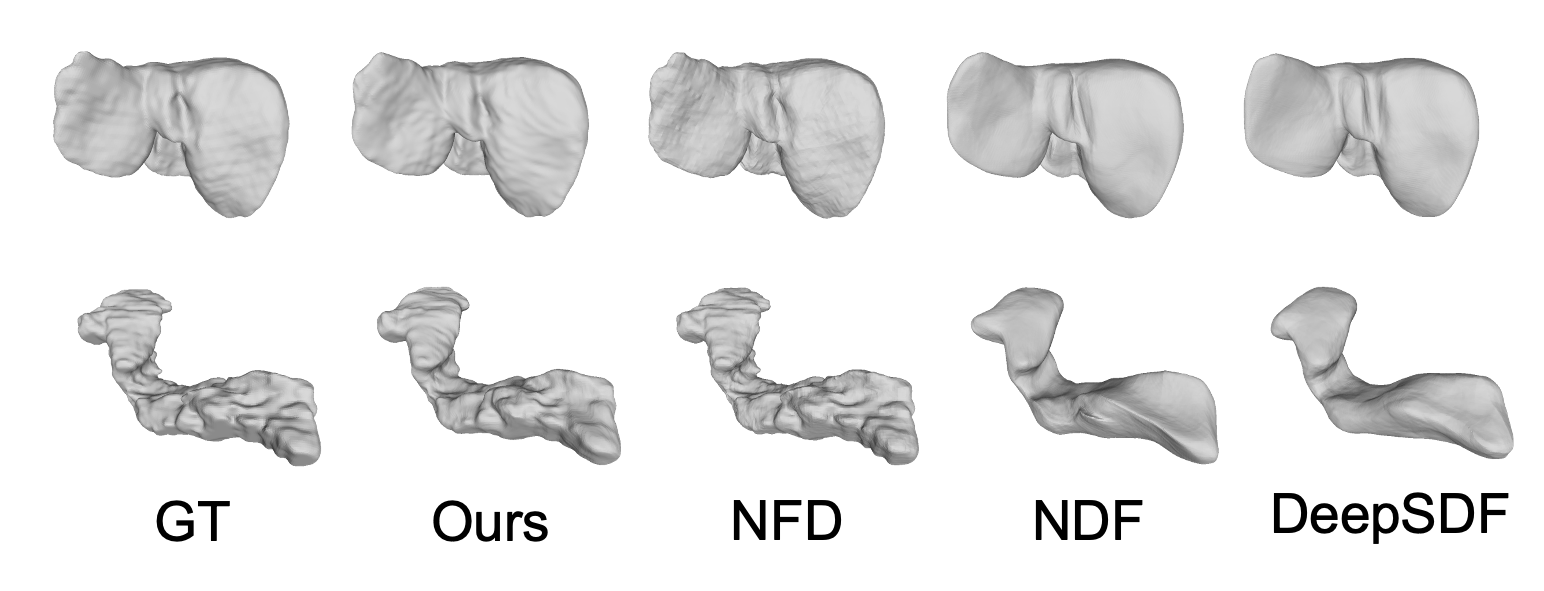}
\vspace{-2em}
\caption{Reconstruction Result on unseend data.}  \label{fig:rec_comp}
\end{figure}

We evaluate our shape representation through two evaluations: \textbf{representation} on training data and \textbf{reconstruction} on unseen data, following the setting of \cite{sun2022topology}. For each point $p$ in the instance space, according to Eq. \ref{eq:ode_forward}, we can get the corresponding destination point $p'$ in the template space, and the trained template module will return the sign distance value for this point. After retrieving the sign distance value for all the grid points, we can then utilize the marching cube algorithm \cite{lorensen1998marching} to extract the mesh for each instance. In the representation comparison, we utilize the trained per-object latent feature to assess the effectiveness of different representation methods. In the reconstruction comparison, we independently optimize the per-object latent feature while keeping the network parameters fixed to evaluate the generability of the methods in shape reconstruction. Fig. \ref{fig:rec_comp} shows the reconstruction results of different methods.

\vspace{-1em}

According to Table \ref{tab:rep}, DIF-Net achieves the best results on the training data representation but worse results on the shape reconstruction tasks, indicating the overfitting on the training data. Our method and NFD achieve similar overall performance, benefiting from the enhanced representation power of the triplane feature. Comparing with NDF, our method achieves superior performance even with a single deformation module, outperforming NDF with 4 consecutive deformation modules. The ablation study conducted on regularization, as shown in Tab. \ref{tab:sup}, demonstrates the significance of our proposed hybrid supervision in achieving accurate reconstruction for new shapes reconstruction.

\begin{table} [ht]
\vspace{-0.5em}
  \begin{center}
    {\small{
\setlength{\tabcolsep}{1mm}{
\begin{tabular}{lcccccc}

\toprule

& \multicolumn{2}{c}{CD Mean($\downarrow$)}  & \multicolumn{2}{c}{NC Mean($\uparrow$)} & \multicolumn{2}{c}{SI Mean($\downarrow$)}   \\
\cmidrule(lr){2-3} \cmidrule(lr){4-5} \cmidrule(lr){6-7}
Model/Data & Pancreas & Liver & Pancreas & Liver & Pancreas & Liver \\
\midrule
DeepSDF & - & - & - & - & - & - \\
NFD & - & - & - & - & - & - \\
\midrule
AtlasNet & 8.08 & 3.46 & 0.703 & 0.823 & 5860 & 29.5 \\
DIT & 0.677 & 0.528 & 0.893 & 0.868 & 346 & 11.8 \\
DIF-Net & 10.5 & 2.06 & 0.694 & 0.832 & 2560 & 4.61 \\
NDF & 0.518 & 0.49 & 0.916 & 0.873 & \textbf{0} & \textbf{2} \\
\midrule
Ours & \textbf{0.099} & \textbf{0.125}  & \textbf{0.946} & \textbf{0.882} & 15 & 8 \\
\bottomrule
\end{tabular}
}}
}
\end{center}
\vspace{-1.5em}

\caption{\textbf{Shape Registration} results on unseen shapes. DeepSDF and NFD don't have scores as they can not capture the point correspondences.}
\vspace{-1.5em}
\label{tab:reg}
\end{table}

\subsection{Point Correspondence and Shape Registration}

As the methods DeepSDF and NFD can only represent the shape without capturing point correspondence, we compare the remaining methods in Table \ref{tab:reg} for shape registration evaluation and the instance shape is represented by deforming the template, as described in Sec. \ref{subsubsec:pc_and_sr}. Following the trajectory defined by the ODE function in Eq. \ref{eq:ode_backward}, each point on the template shape moves towards the corresponding destination point on the instance space. As a result, the instance shape, defined as the deformed template, shares the same underlying topology as the template shape, ensuring consistent connectivity. The diffeomorphic deformation from the template towards instance shapes is shown in the left half of Fig. 1. 

\begin{table} [ht]
  \begin{center}
    {\small{
\setlength{\tabcolsep}{1mm}{
\begin{tabular}{lcccccc}

\toprule

& \multicolumn{2}{c}{FID Mean($\downarrow$)}  & \multicolumn{2}{c}{Prec. Mean($\uparrow$)} & \multicolumn{2}{c}{Recall Mean($\uparrow$)}   \\
\cmidrule(lr){2-3} \cmidrule(lr){4-5} \cmidrule(lr){6-7}
Model/Data & Pancreas & Liver & Pancreas & Liver & Pancreas & Liver \\
\midrule
DeepSDF $\triangle$  & 99.46 & 93.74 & 0.810 & 0.858 & 0.078 & 0.089\\ 
DeepSDF $\bigstar$  & 80.03 & 85.64 & 0.729 & 0.810 & 0.430 & 0.534 \\ 
NDF $\triangle$  & 69.66 & 60.50 & 0.797 & 0.714 & 0.508 & 0.593 \\ 
NDF $\bigstar$  & 69.66 & 66.45 & 0.844 & 0.821 & 0.505 & 0.571 \\ 
\midrule
PVD & 89.26 & 86.32 & 0.760 & 0.821 & 0.420  & 0.466 \\ 
3D-LDM  & 78.64 & 79.58 & 0.782 & 0.824 & 0.470 & 0.554 \\ 
NFD  & 72.83 & 74.24 & 0.812 & 0.831 & 0.523 & 0.560 \\ 
\midrule
Ours & \textbf{52.01} & \textbf{48.54} & \textbf{0.992} & \textbf{0.994} & \textbf{0.661} & \textbf{0.613} \\ 
\bottomrule
\end{tabular}
}}
}
\end{center}
\vspace{-1.5em}

\caption{\textbf{Shape generation} results. Our method achieve better performance according to the FID, precision and recall. $\triangle$ denotes sampling from Gaussian distribution while $\bigstar$ denotes sampling from PCA.}
\vspace{-1.5em}
\label{tab:gen}
\end{table}

To evaluate the point correspondence and shape registration results, we compare the deformed template with the corresponding ground truth instance shape. We also utilize self-intersection as a metric to assess the preservation of topology and geometric fidelity during the deformation. To ensure a fair comparison, we remesh the template meshes to have the same number of vertices (5000), following the approach in \cite{sun2022topology}. Based on the comparison presented in Table \ref{tab:reg}, our proposed method achieves better registration accuracy and correct dense correspondence, with only slight self-intersection, which can be considered negligible given the large number of vertices and faces in the template shape.

\subsection{Shape Generation}
\vspace{-1.5em}
\begin{figure}[ht]
\includegraphics[width=1\linewidth]{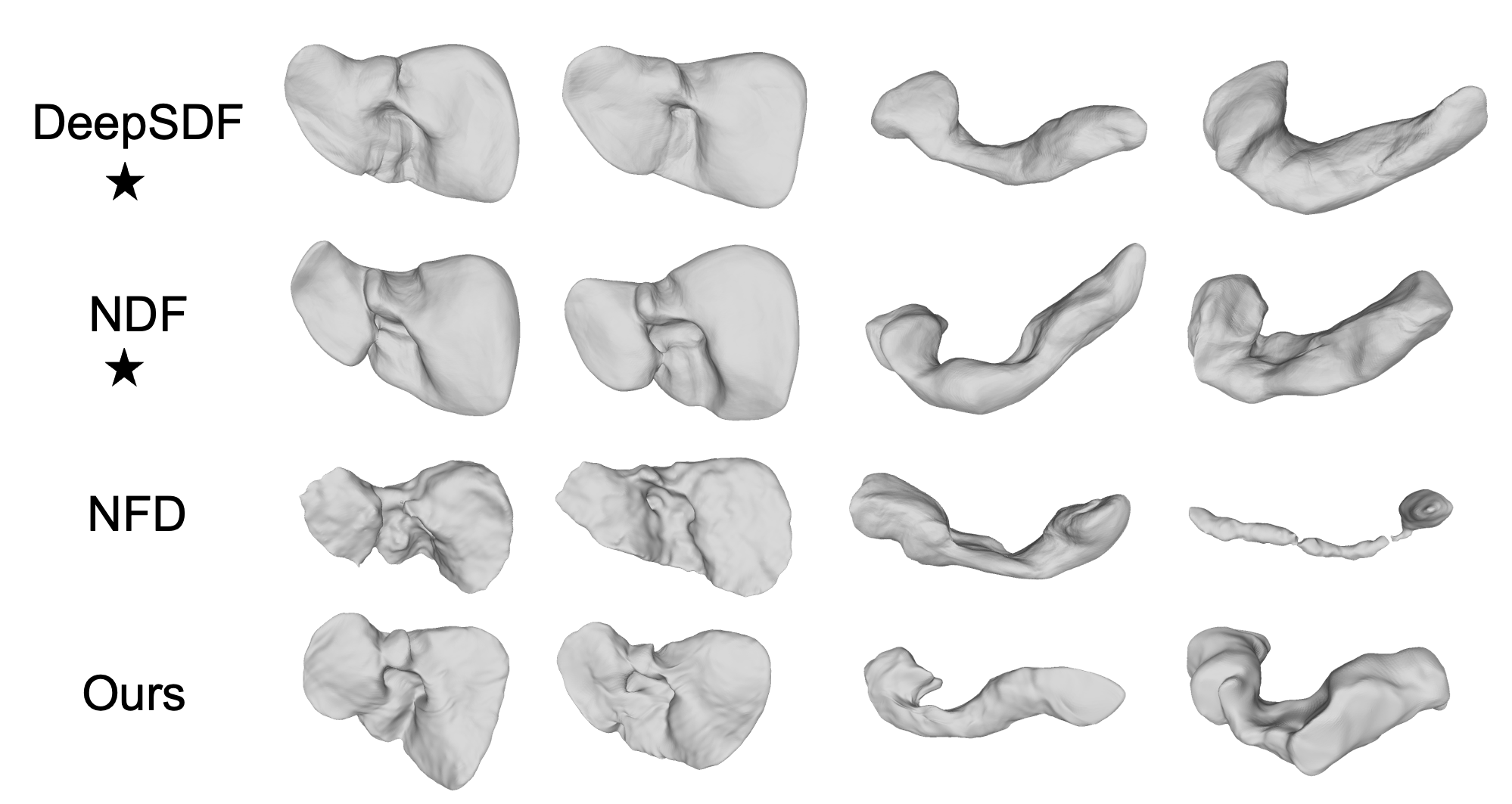}
\vspace{-2em}
\caption{Visualization of generated 3D shape.}  \label{fig:vis_gen}
\vspace{-1em}
\end{figure}

Table \ref{tab:gen} presents the evaluation of shape generation across different methods. For DeepSDF and NDF, we sample global latent vectors from a Gaussian distribution and perform PCA analysis, where the parameters are determined by grid search. However, similar to the results in previous experiments, the shapes sampled from DeepSDF and NDF tend to be smoother compared to real instance shapes. 
PVD is capable of generating variable shapes, but it is limited by its nature to generate only coarse object shapes. 3D-LDM attempts to capture the distribution of the global latent vectors of DeepSDF, but still faces the smoothing issue from the global latent vector. NFD can also generate variable shapes. However, compared to our methods, the shapes generated by NFD may not preserve topology, resulting in potentially separated components in the generated shapes, as shown in Fig. \ref{fig:vis_gen}. In contrast, our method focuses on generating diffeomorphic deformations encoded by triplane features. The new shapes are generated by deforming the template, allowing us to achieve high fidelity and variability while preserving the underlying topology.

\subsection{Ablation Study}

\noindent\textbf{Supervision} 
Table \ref{tab:sup} highlights the significance of our global supervision in shape reconstruction, mitigating the risk of local minima. While incorporating additional mesh supervision improved the results marginally, it also increased computational and memory demands. Thus, we opted to utilize global supervision in our approach.
\begin{table} [ht]
\vspace{-0.7em}
  \begin{center}
    {\small{
\setlength{\tabcolsep}{2.2mm}{

\begin{tabular}{lcccccc}

\toprule

& \multicolumn{2}{c}{CD Mean($\downarrow$)}  & \multicolumn{2}{c}{NC Mean($\uparrow$)}  \\
\cmidrule(lr){2-3} \cmidrule(lr){4-5}
Model/Data & Pancreas & Liver & Pancreas & Liver \\
\midrule
Ours & \textbf{0.082} & 0.116 & \textbf{0.961} & 0.885\\
Ours - Global Sup.  & 0.264 & 0.368 & 0.932 & 0.877 \\
Ours + Mesh Sup.  & \textbf{0.082} & \textbf{0.112}  & 0.960 & \textbf{0.886} \\
\bottomrule
\end{tabular}
}}
}
\end{center}
\vspace{-1.5em}

\caption{Shape Reconstruction with various supervision. }
\vspace{-1em}
\label{tab:sup}
\end{table}

\noindent\textbf{Feature Representation} We explored the use of 3D voxel-grid features as an alternative to triplane features, and found that they yielded similar results as shown in Table \ref{tab:loc}. However, voxel-grid features required more computation and memory resources for representation and generation tasks. In contrast, triplane feature representation achieved high reconstruction accuracy with improved memory and computation efficiency.
\begin{table} [ht]
\vspace{-0.7em}
  \begin{center}
    {\small{
\setlength{\tabcolsep}{3mm}{
\begin{tabular}{lcccccc}

\toprule

& \multicolumn{2}{c}{CD Mean($\downarrow$)}  & \multicolumn{2}{c}{NC Mean($\uparrow$)}  \\
\cmidrule(lr){2-3} \cmidrule(lr){4-5}
Model/Data & Pancreas & Liver & Pancreas & Liver \\
\midrule
Vector & 0.512 & 0.476 & 0.917 & 0.873 \\
Triplane & \textbf{0.082} & 0.116 & \textbf{0.961} & \textbf{0.885} \\
Voxel & 0.146 & \textbf{0.112} & 0.957 & \textbf{0.885} \\

\bottomrule
\end{tabular}
}}
}
\end{center}
\vspace{-1.5em}

\caption{Shape Reconstruction with various feature representations. }
\vspace{-1.5em}
\label{tab:loc}
\end{table}

\section{Conclusion}

In this paper, we introduce Hybrid Neural Diffeomorphic Flow (HNDF) as a novel approach for topology-preserving shape representation and generation. Our method leverages the expressive power of triplane representation, enabling accurate dense correspondence and high representation accuracy. The proposed hybrid supervision plays a crucial role in capturing both local and global correspondence. Unlike existing methods that primarily focus on directly generating shapes, we explore the concept of generating shapes using deformed templates to preserve the underpying topology. We present a baseline method for topology-preserving shape generation and will continue our exploration for more complex shapes and scenarios. By presenting our research, we aim to contribute to the 3D vision community and provide insights into the potential of topology-preserving shape representation and generation.

\newpage
{\small
\bibliographystyle{ieee_fullname}
\bibliography{egbib}
}

\end{document}


\title{Hybrid Neural Diffeomorphic Flow for Shape Representation and Generation via Triplane}  

\maketitle
\thispagestyle{empty}
\appendix

\section{Data Source and Data Preparation}
For the Pancreas-CT \cite{roth2015deeporgan} and Inhouse Liver \cite{chen2021deep} datasets, we follow the same data splitting strategy as described in \cite{sun2022topology}. For the Pancreas-CT dataset, 61 out of 81 cases are used for training, while the remaining 21 cases are used for testing the generalization capability in the reconstruction tasks. As for the Inhouse Liver dataset, we use 145 instances for training and 45 instances for testing.

For the generation tasks, we utilize the Abdomen1k dataset \cite{ma2021abdomenct}, which contains 573 valid liver instances and 693 valid pancreas instances after filtering out incomplete shapes or shapes with large spacings. For our methods, DeepSDF, NDF and NFD, we use 300 cases for each organ to train the deformation module and template module. After training, we perform reconstruction for all shapes to obtain the training data for the diffusion models. For other generative models, we use all available data to train the models.

The initial data format for each instance is a mask. We start by extracting the mesh from the ground truth mask using the marching cubes algorithm \cite{lorensen1998marching}, following the approach in \cite{sun2022topology}. To improve the mesh quality, we apply laplacian smoothing to remove artifacts. Once the mesh is obtained, we sample the signed distance values. Specifically, we uniformly sample 20 percent of the values throughout the entire space, while the remaining values are sampled near the surface of the mesh.

\section{Training and Inference Settings}
Each shape in our approach is represented by triplane features of size $L \times L \times 3C$, where $L$ represents the dimensions (in our case, $L = 96$) and $C$ represents the number of channels (in our case, $C = 4$). These triplane features are optimized during both the training and inference stages of our model.

During training, we use an initial learning rate of 0.005 for the deformation module $\mathcal{D}$ and the template module $\mathcal{T}$, while the initial learning rate for the triplane feature is set to 0.001. After every 500 epochs, all learning rates are multiplied by a factor of 0.5. We optimize the parameters using the Adam optimizer.
For reconstruction of unseen shapes, we fix the deformation module $\mathcal{D}$ and the template module $\mathcal{T}$, and only optimize the per-object triplane features with 1600 iterations. We use an initial learning rate of 0.0005 and then halve the learning rate every 800 iterations.

\begin{figure}[ht]
\centering
\includegraphics[width=0.9\linewidth]{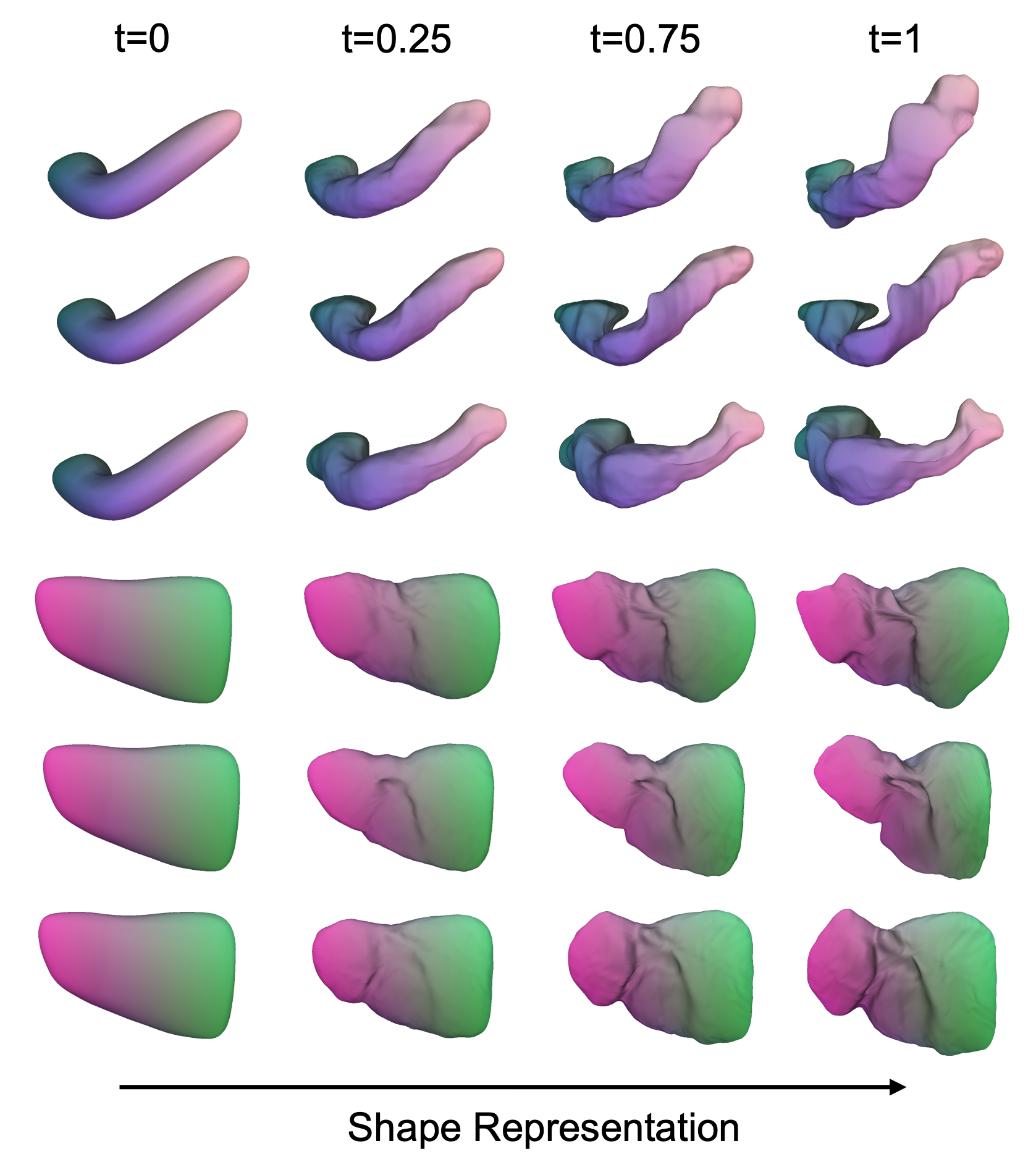}
\caption{Additional \textbf{Shape Reconstruction} result on unseend data.}  \label{fig:supp_rep}
\end{figure}

\begin{figure}[ht]
\centering
\includegraphics[width=0.9\linewidth]{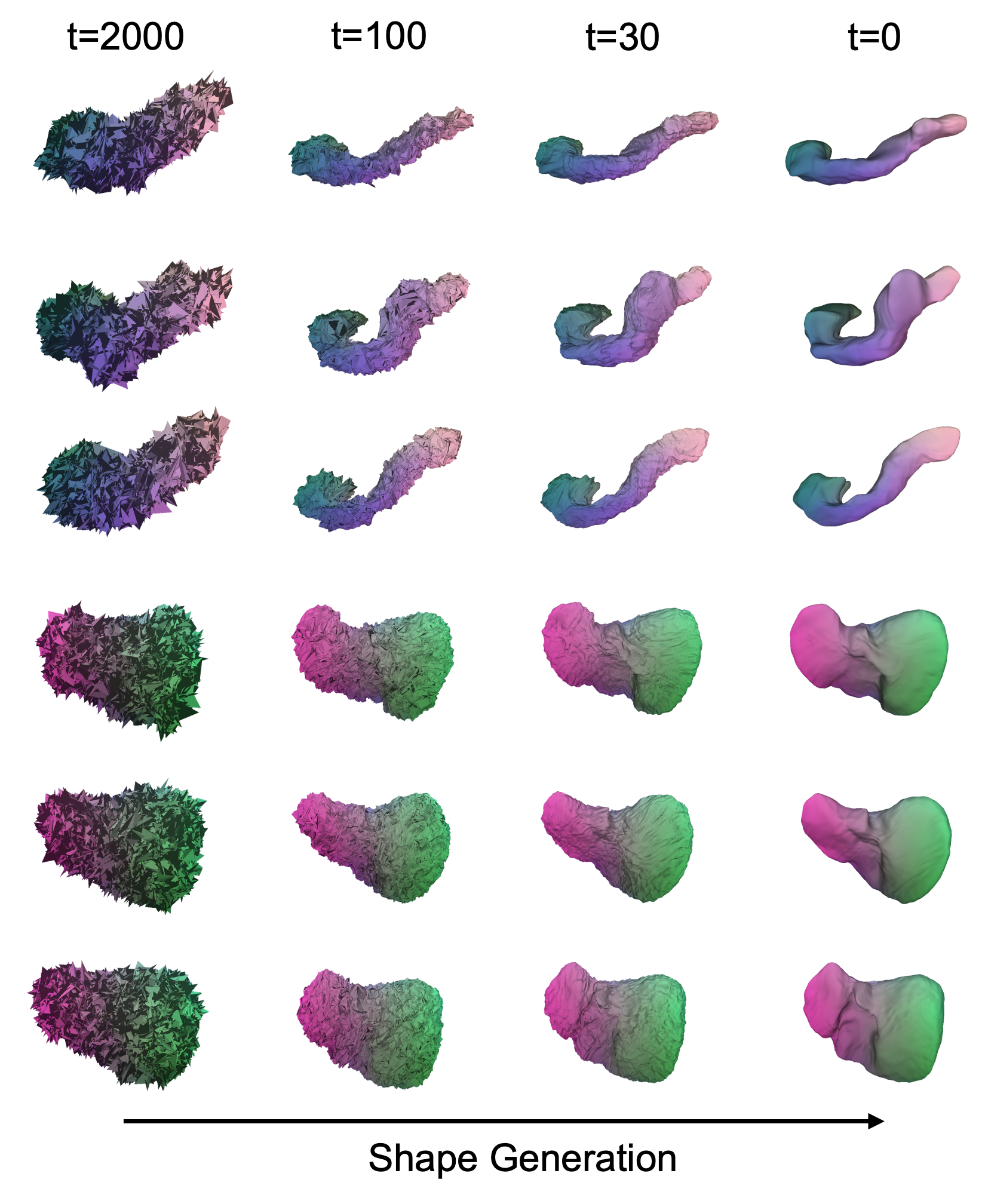}
\caption{Additional \textbf{Shape Generation} results from generated triplane features.}  \label{fig:supp_gen}
\end{figure}


\newpage
{\small
\bibliographystyle{ieee_fullname}
\bibliography{egbib}
}